\documentclass[letterpaper, 10 pt, conference]{ieeeconf}  % Comment this line out if you need a4paper

\IEEEoverridecommandlockouts                              % This command is only needed if 
\usepackage{cite}
\usepackage{amsmath,amssymb,amsfonts}
\usepackage{algorithmic,algorithm,multirow,threeparttable}
\usepackage{graphicx}
\usepackage{stfloats}
\usepackage{booktabs} % 用于三线表格
\usepackage{tabularx} % 用于自动调整列宽
\usepackage{textcomp}
\usepackage{xcolor}
\usepackage[caption=false]{subfig}
\usepackage{bm}
\title{\LARGE \bf Multi-Arm Robot Task Planning for Fruit Harvesting Using Multi-Agent Reinforcement Learning}

\author{Tao Li$^{1}$, Feng Xie$^{2}$, Ya Xiong$^{1}$ and Qingchun Feng$^{1*}$% <-this % stops a space
	\thanks{*This work was funded by Beijing Science and Technology Plan Project of China under Grant (Z201100008020009), Youth Research Foundation of Beijing Academy of Agriculture and Forestry Sciences (QNJJ202318), Beijing Nova Program (20220484023)}% <-this % stops a space
	\thanks{$^{1}$T. Li, Y. Xiong and Q. Feng are with Intelligent Equipment Research Center, Beijing Academy of Agriculture and Forestry Sciences, Beijing 100097, China. Q. Feng is the corresponding author. E-mail: {\tt\small \{lit, yaxiong and fengqc\}@nercita.org.cn}}%
	\thanks{$^{2}$F. Xie is with School of Agricultural Engineering, Jiangsu University, Zhenjiang 212013, China. Email: {\tt\small 2111916009@stmail.ujs.edu.cn}}%
}

\begin{document}
	\maketitle
	\thispagestyle{empty}
	\pagestyle{empty}

	\begin{abstract}
The emergence of harvesting robotics offers a promising solution to the issue of limited agricultural labor resources and the increasing demand for fruits. Despite notable advancements in the field of harvesting robotics, the utilization of such technology in orchards is still limited. The key challenge is to improve operational efficiency. Taking into account inner-arm conflicts, couplings of DoFs, and dynamic tasks, we propose a task planning strategy for a harvesting robot with four arms in this paper. The proposed method employs a Markov game framework to formulate the four-arm robotic harvesting task, which avoids the computational complexity of solving an NP-hard scheduling problem. Furthermore, a multi-agent reinforcement learning (MARL) structure with a fully centralized collaboration protocol is used to train a MARL-based task planning network. Several simulations and orchard experiments are conducted to validate the effectiveness of the proposed method for a multi-arm harvesting robot in comparison with the existing method.
	\end{abstract}

	\section{Introduction}
The labor-intensive process of harvesting incurs high costs in fruit production\cite{baoling2019}. To reduce labor costs, the urgent need to address the bottleneck of the fruit industry is an autonomous and unmanned harvest. Over the last decade, there has been an increase in the use of robotic harvesters, which are viewed as a promising technique.

With recent advances in computer and sensor technology, harvesting robots in orchards have received extensive attention, and considerable progress has been made in fruit detection and localization\cite{gongal2015sensors}, path planning\cite{2020Path}, mechanical design\cite{jin2021development}, etc. In recent years, harvesting robots have been developed for apples\cite{kootstra2021selective}, oranges\cite{zhou2022intelligent}, tomatoes\cite{rong2022fruit}, pineapples\cite{kurbah2022design}, among others.

Recently, there has been an increasing number of studies on multi-arm harvesting robots \cite{williams2019robotic, xiong2020autonomous, 2019Dual} to enhance operational efficiency, resulting in significant advances, such as a dual-arm strawberry harvesting robot \cite{xiong2020autonomous}, 12-arm apple harvesting robot \cite{FFR}, 4-arm kiwifruit harvesting robot\cite{williams2019robotic}, etc. Some multi-arm robots have been successfully commercialized.
A multi-arm harvesting robot typically comprises several robotic arms, grippers, and vision units to simultaneously perform harvesting operations, thereby increasing efficiency.

\begin{figure}[t]
	\centering
	\includegraphics[width=1\linewidth]{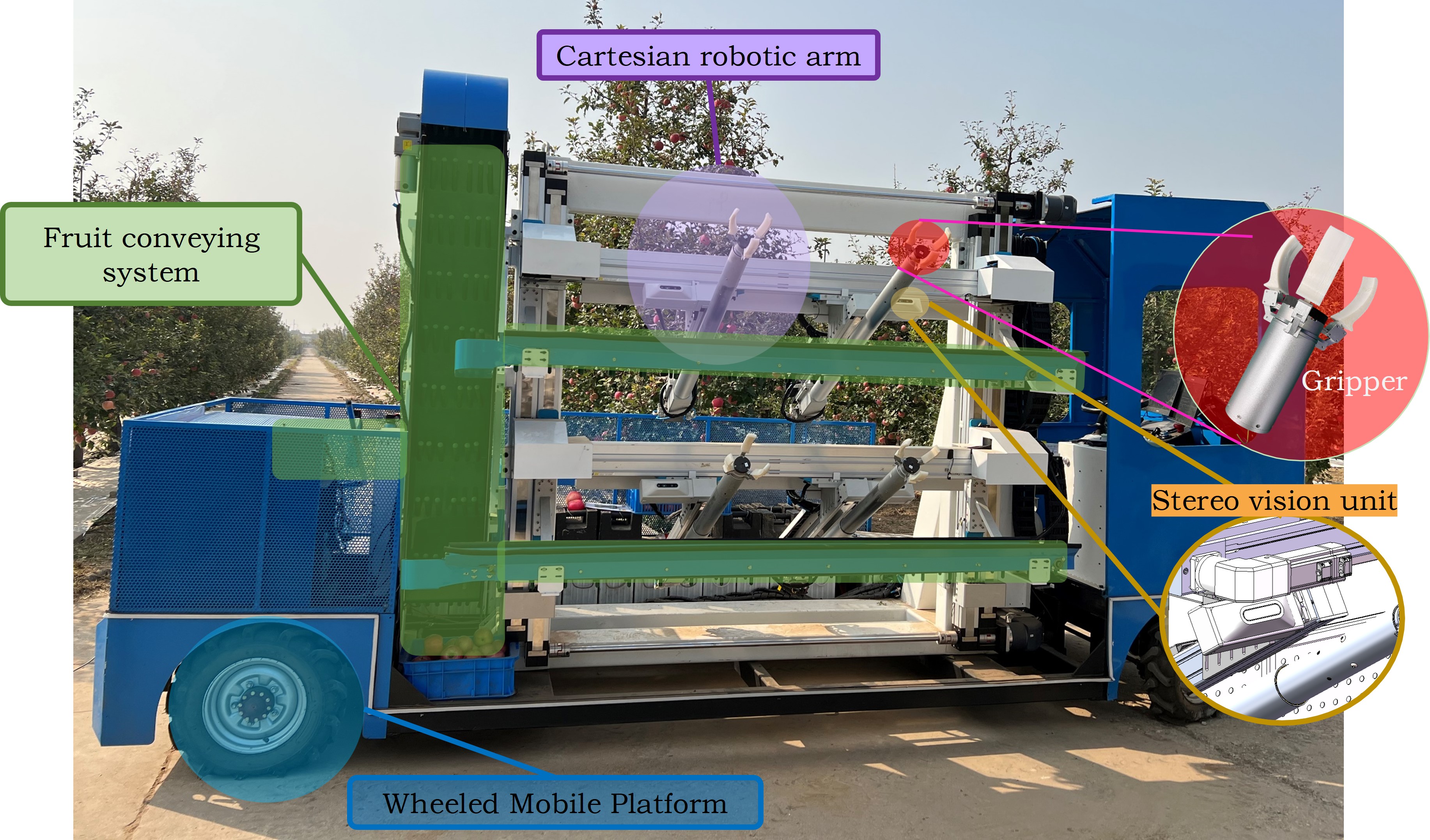}
	\caption{The multi-arm apple harvesting robot presented in this work.}
	\label{fig:robot-pic}
\end{figure}
Despite the advantages of a large workspace and scale of perception, multi-arm robots face several technical challenges. Generally speaking, multi-arm robots require more complicated task planning strategies than single-arm robots to achieve better collaboration and efficiency\cite{li2022advance}.
The task planning for multi-arm harvesting robots is two-fold: picking sequence and fruit target allocation.
When a robot needs to pick a large number of fruits, the order in which it picks them can result in arms having to travel different distances between targets \cite{kurtser2020planning}. If the distances are large, the robot may end up taking unnecessary paths while moving from one fruit to another.
Second, the allocation of fruit targets means that each arm knows which target to pick, especially for some targets that can be reached by other arms \cite{bac2014harvesting}. It is significant for multi-arm collaboration to avoid inter-arm conflicts that usually cause extra waiting time. All arms have tasks to do only when they have their own targets in their working zones. If fruits are not evenly distributed \cite{li2016characterizing}, one or more arms could be vacant until the others finish their tasks.
Consequently, task planning is necessary for a multi-arm harvesting robot to reduce redundancy motions and conflicts among arms, thereby maximizing working performance.

However, the task planning for multi-arm robots still faces many technical challenges. 
In practical operation, fruit picking is often not successful in one attempt \cite{wang2022geometry}, and there is a possibility of failure. In order to improve the harvesting completion rate, the robot needs to have the ability to re-grasp failed fruits, instead of mechanically executing its original task planning strategy. However, existing heuristic planning methods have large computational complexity and are difficult to meet real-time control requirements. 
Another challenge is the modelling and solving the time optimization model for multi-arm picking of multiple fruits. 
The robot in this study involves multiple constraints such as mechanical structure interference, joint degree of freedom coupling, and multi-arm interaction. Presently, there is a scarcity of viable solutions aimed at elucidating the distinct impact of decision-making processes on overall operational time, alongside the modelling and solving the optimization problem that effectively incorporates multiple constraints.

To address the problems, we develop a task planning scheme for multi-arm harvesting robots. The contributions of this work are as follows:
\begin{enumerate}
\item We present a harvesting robotic system that consists of four Cartesian arms, cameras, a conveying system, illuminating devices, and a mobile platform, as shown in Figure \ref{fig:robot-pic}.
\item To minimize the global harvesting cost of time for multiple arms, we use a Markov game to model the sequential decision-making problem.
\item To solve the multi-arm cooperation problem, we propose a fully centralized collaboration strategy of multi-agent reinforcement learning (MARL) framework.
\end{enumerate}

\section{Related work}
Various approaches have been proposed to enable cooperative fruit harvesting with multi-arm robots. Based on the literature, multi-arm robot collaboration can be classified into three categories. The first category is quasi-human dual-arm, wherein one arm performs the picking action while the other assists \cite{ling2019dual, sepulveda2020robotic}. This mode is suitable for skill-intensive harvesting tasks involving slender stalks, such as those of eggplants and sweet peppers, where it is not appropriate to pull and separate fruits during harvesting. Instead, the end effector must accurately position the fruit stalk for cutting. In this mode, robots can remove obstructions to expose the fruit stems and detach a fruit, thereby facilitating the harvesting process. Fortunately, growers are increasingly manipulating plant form or shape to encourage production. For instance, pruning into a planar canopy can improve fruit accessibility, and apple growers in the US have been removing fruits adjacent to trellis wires and trunks to optimize fruit distribution for robot harvesting \cite{hohimer2019design}.

To further enhance robot harvest efficiency, many multi-arm robots employ the parallel mode of collaboration, which is the second category, as exemplified by FFR\cite{FFR}, AGROBOT\cite{bogue2020fruit}, Harvest CROO\cite{delbridge2021robotic}, and the robots in \cite{xiong2020autonomous, williams2020improvements}. In this mode, arms pick fruits with spatial isolation and shorten the average cycle time of single fruit harvest. For example, Harvest CROO's harvester claims to pick three fruits every ten seconds. It is worth noting that the definitions of cycle time, a key indicator of harvesting efficiency, vary among researchers \cite{zhou2022intelligent}. In some studies, time of perception or inter-fruit traversal is neglected to achieve a shorter cycle time, but the overall picking rate could be lower in a large-scale harvesting operation\cite{ringdahl2019evaluation}. The reasons are twofold. First, the traversing time could be large, which means that robots have redundant paths during an arm traversing from one fruit to another. Second, arm vacancy matters. All arms have tasks to do only when they have their targets in their working zones. If fruits are not evenly distributed, one or more arms could be vacant until the others finish the tasks. In this case, the cycle time depends on the uniformity of fruit distribution.

To further improve the productivity of robotic harvesters, a cooperative mode of collaboration is required, which is the third category. In this type of collaboration, there is overlapping workspace among the arms, and cooperative picking needs to be implemented in the overlapping space to avoid interference and improve efficiency. This is different from both multi-arm parallel and dual-arm cooperative methods, and combines the above two characteristics. The focus is on how to arrange the traversal order of randomly distributed multiple target fruits in a reasonable way to maximize picking efficiency.
Barnett et al. investigated a multi-arm kiwifruit harvesting robot by considering task partitioning and reachability to achieve the uniformity of fruit distribution and minimize task completion time \cite{barnett2020work,au2020workspace}. Mann et al. studied an optimal robot in terms of the number of arms, manipulator capabilities, and robot speed, allocating tasks to each manipulator to yield the maximum harvest \cite{mann2016combinatorial}. Li et al. presented a method of task planning for multi-arm apple harvesting robots to plan the picking sequence for each arm to reduce arm operation vacancy and collision, claiming that the harvesting efficiency could be increased by up to 4.25 times \cite{li2021task}.

\section{Multi-arm apple harvesting robot}\label{sec:2}

The multi-arm harvesting robot investigated in this work is designed for standard apple orchards with simple, narrow, accessible, and productive (SNAP) fruiting-wall canopy architecture, as shown in Figure \ref{fig:robot-pic}. It comprises a wheeled mobile platform, four Cartesian robotic arms, stereo vision units, a fruit conveying system, a control system, and a power system.
The visual sensing system includes 4 RGBD cameras that acquire visual information of target fruits, which is then transmitted to an edge computing module for fruit localization.
The control system is for computation, image processing, and human-machine interaction, including a computer, two edge computing devices, and a screen. An operator can monitor the robot's operating state, such as the number of fruits picked, operation time, battery usage, and fault diagnosis.
The conveying system includes fruit storage boxes and three conveyors that gather the fruits delivered by the upper and lower conveyors to the fruit storage box.
\begin{figure}[h]
	\centering
	\subfloat[Linear guides and arms]{\includegraphics[width=0.5\linewidth]{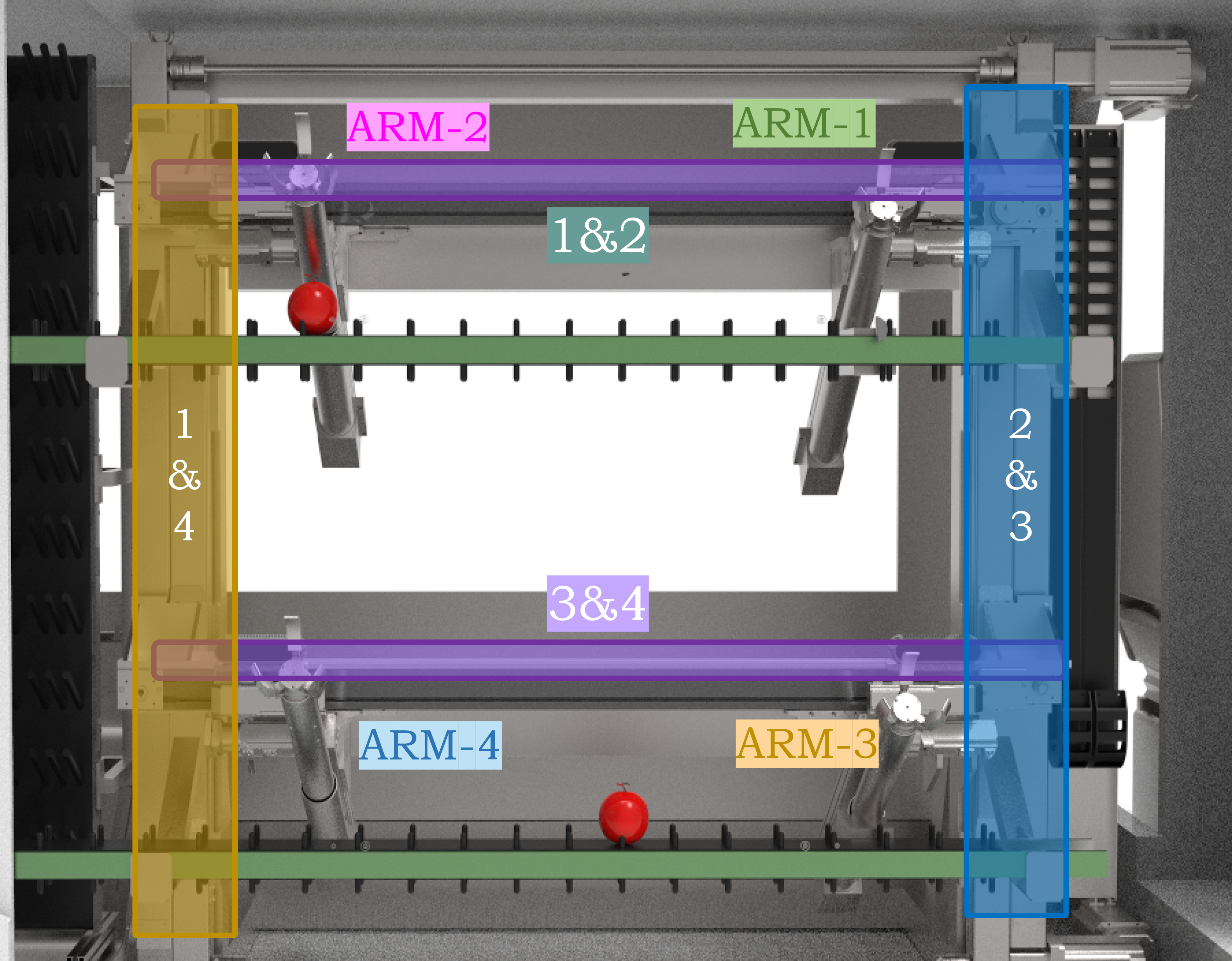}}
	\subfloat[Workspace divisions]{\includegraphics[width=0.5\linewidth]{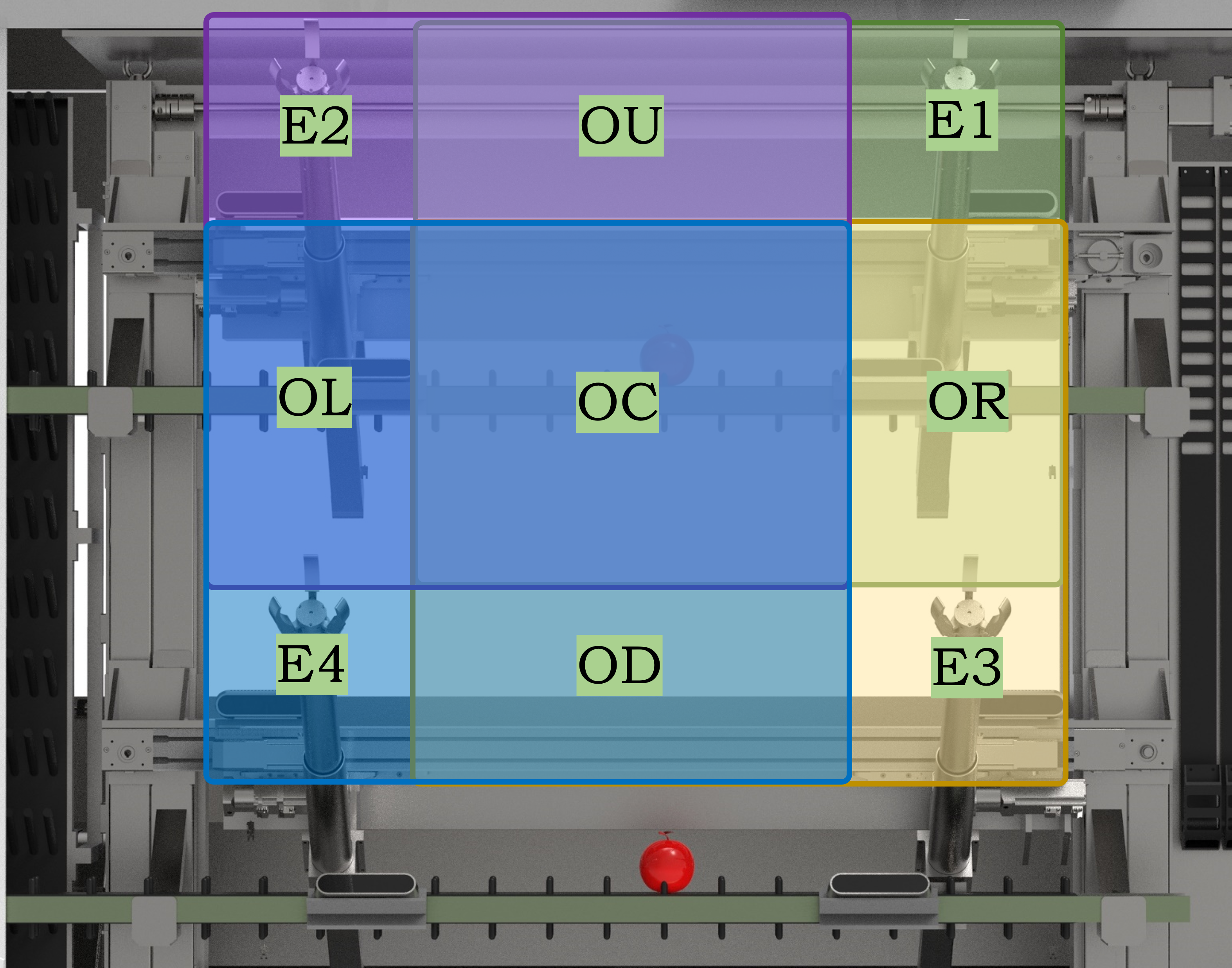}}
	\caption{Schematic diagram of the proposed multi-arm mechanical structure and workspace of the harvesting robot.}
	\label{fig:workingzone}
\end{figure}
The multi-arm system's overall structure is presented in Figure \ref{fig:workingzone}. The arms' workspaces overlap in the shared linear guides, forming divisions illustrated in Figure \ref{fig:workingzone}(a). Figure \ref{fig:workingzone}(a) depicts a four-arm diagram (numbered 1 to 4) and the shared guides, numbered 1\&4, 2\&3, 1\&2, and 3\&4. Figure \ref{fig:workingzone}(b) illustrates the workspace profile divided into nine sections, including exclusive areas E1 to E4, which only allow a single arm to enter, and common areas OU to OC, which allow multiple arms to enter.

\begin{figure}
	\centering
	\includegraphics[width=.8\linewidth]{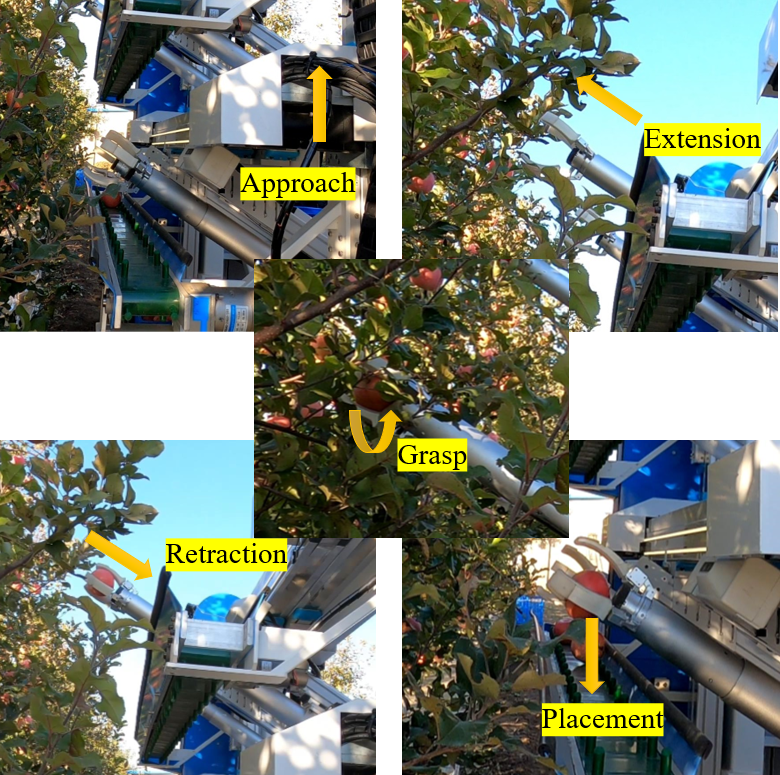}
	\caption{Five sequential actions of robotic arms in harvesting operation.}
	\label{fig:actions}
\end{figure}

For the purpose of analysis, we define the joint-1 as the vertical movement joint, which allows the robot to move up and down. Joint-2 is defined as the horizontal movement joint, enabling the robot to move from side to side. Finally, joint-3 is defined as the longitudinal movement joint, allowing the robot to move forward and backward. These definitions provide clear and precise names for each joint, which can be used consistently throughout our analysis. Based on the mechanical structure of the robot, No.1 and No.2 are grouped as group-U, and No.3 and No.4 are grouped as group-D. Joint-1 is the coupling joint that synchronizes the operations of the inner-group arms. In this work, the harvesting actions of the robotic arms are sequential, as shown in Figure \ref{fig:actions}, including approach (A), extension (E), grasp (G), retraction (R), and placement (P). During harvesting, the actions of approach (A), extension (E), and grasp (G) occupy joint-1, while the actions of retraction (R) and placement (P) release it. When an arm performs R and P, its inner-group neighbors can perform A-E-G simultaneously.

Moreover, during the picking process, fruit targets may be attempted multiple times to ensure the harvesting rate, and adjacent robotic arms cannot access targets that are too close to each other due to mechanical limitations.
Under the aforementioned constraints, the goal of the planning is to find a picking sequence and a fruit target allocation scheme for each arm to optimize the overall operating time. The overall operating time is determined by the last arm that finishes its task.
In summary, the objective of this study is to find a picking sequence and allocation scheme for the four arms to optimize the operation time, based on the above constraints.
\section{Multi-agent reinforcement learning based Multi-arm task planning}
\subsection{Markov Game}
Markov Decision Processes (MDPs)\cite{puterman1990markov} are the classical representation of sequential decision making. At each step $t$ of an MDP, the system's next state can be computed based on the current state and action, defined by a tuple $(\mathbf{S}, \mathbf{A}, f, g, \gamma)$. Here, $\mathbf{S}$ represents the set of system states, $\mathbf{A}$ is the set of allowable actions, $f:s_k\times a_k\rightarrow s_{k+1}$ is the state transition function, $g:s_k\times a_k\rightarrow r_k$ is the reward function, and $\gamma\in [0,1]$ is the discount factor compensating for the effect of immediate and future rewards.

However, when multiple agents are involved, the environment's state dynamics are influenced by other agents, and MDPs are no longer adequate for modeling such problems. In this case, the framework of Markov games (MGs) is used, which is formulated by a tuple  $(N, \mathbf{S}, \{\mathbf{A}^i\}_{i\in N}, f, \{g^i\}_{i\in N}, \gamma)$, where $N=1,2,3,\ldots,N$ is the set of $N>1$ agents; $\mathbf{S}, f$ and $\gamma$ have the same definitions as in MDPs; $\{\mathbf{A}^i\}_{i\in N}$ denotes the action space of the $i$-th agent; $\mathbf{A}:=A^1\times A^2\times A^3\times \cdots A^N$ is the joint action space. The reward function $\{g^i\}_{i\in N}$ represents the instantaneous reward obtained by the $i$-th agent.

In the present study, we consider a MG with two agents, namely group-U and group-D, controlling four arms divided into two groups. The system state $\mathbf{S}$, the action sets $\{\mathbf{A}^i\}_{i\in N}$, and the reward function $\{g^i\}_{i\in N}$ will be introduced in the following subsections.
\subsection{System State}
At time $k$, the states of the task planning decision model presented can be represented by the set $\mathbf{s}_k = {\mathbf{D}_k,\mathbf{P}_k,\mathbf{\Psi}_k,\mathbf{\Phi}_k,\mathbf{O}_k}$. Here, $\mathbf{s}_k\in \mathbf{S}$, $\mathbf{D}_k \in \mathbb{R}^{N\times 3}$ represents the fruit distribution, $\mathbf{P}_k\in \mathbb{R}^{M\time 4}$ represents the arm positions and states, where the coordinates of the arms are given with respect to the base frame, and the arm state is defined as either 0 (at AEG) or 1 (at RP). $\mathbf{\Psi}_k \in \mathbb{N}^{N\times M}$ represents the allocation states, where the $m$-th element in the $n$-th row indicates whether the $n$-th fruit is assigned to the $m$-th arm (0: "no", 1: "yes"). $\mathbf{\Phi}_k \in \mathbb{N}^{N\times M}$ represents the number of attempts, where the $m$-th element in the $n$-th row indicates the number of times the $n$-th fruit has been attempted to be picked by the $m$-th arm (0: "not picked yet", 1-3: number of attempts). Finally, $\mathbf{O}_k\in \mathbb{N}^{N\times M}$ represents whether the target fruit has been picked, where the $m$-th element in the $n$-th row indicates whether the $n$-th fruit has been picked by the $m$-th arm (0: "no", 1: "yes").
\begin{figure*}[b]
	\centering
	\includegraphics[width=.8\linewidth]{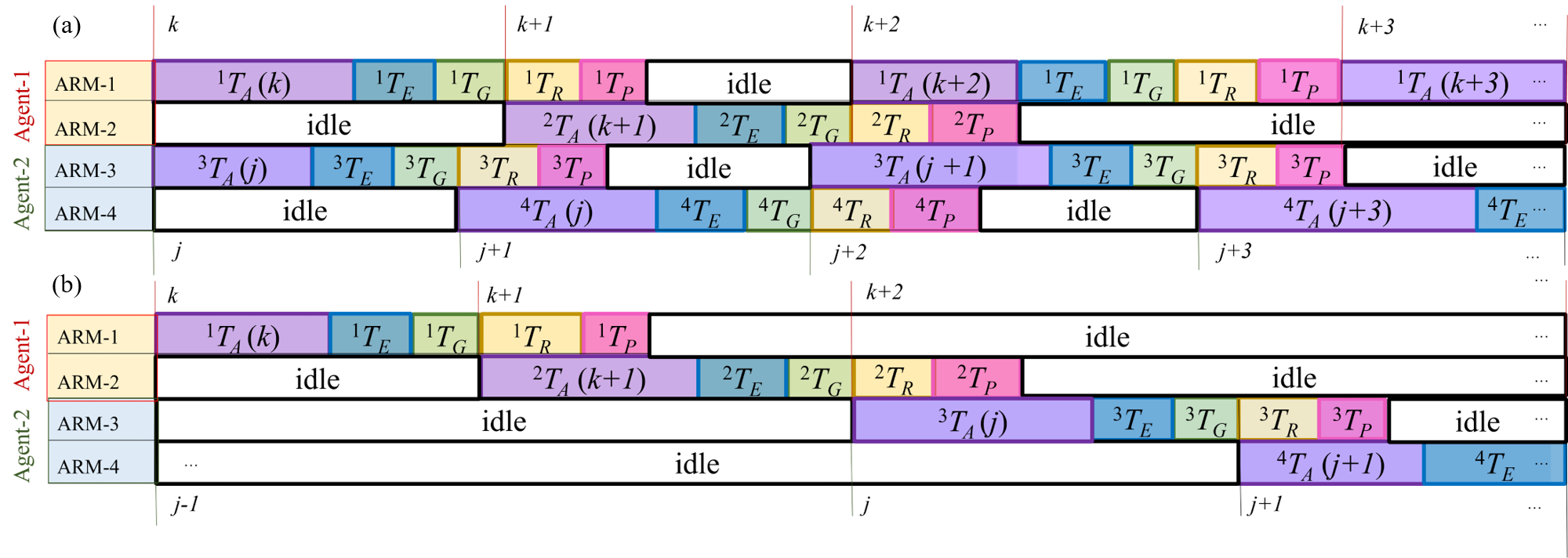}
	\caption{State transitions of multi-arm cooperation.(a) shows transitions in the case of parallel operation; (b) illustrates transitions in the case of alternation.}
	\label{fig:timesequence}
\end{figure*}
\begin{figure*}[b]
	\centering
	\includegraphics[width=\linewidth]{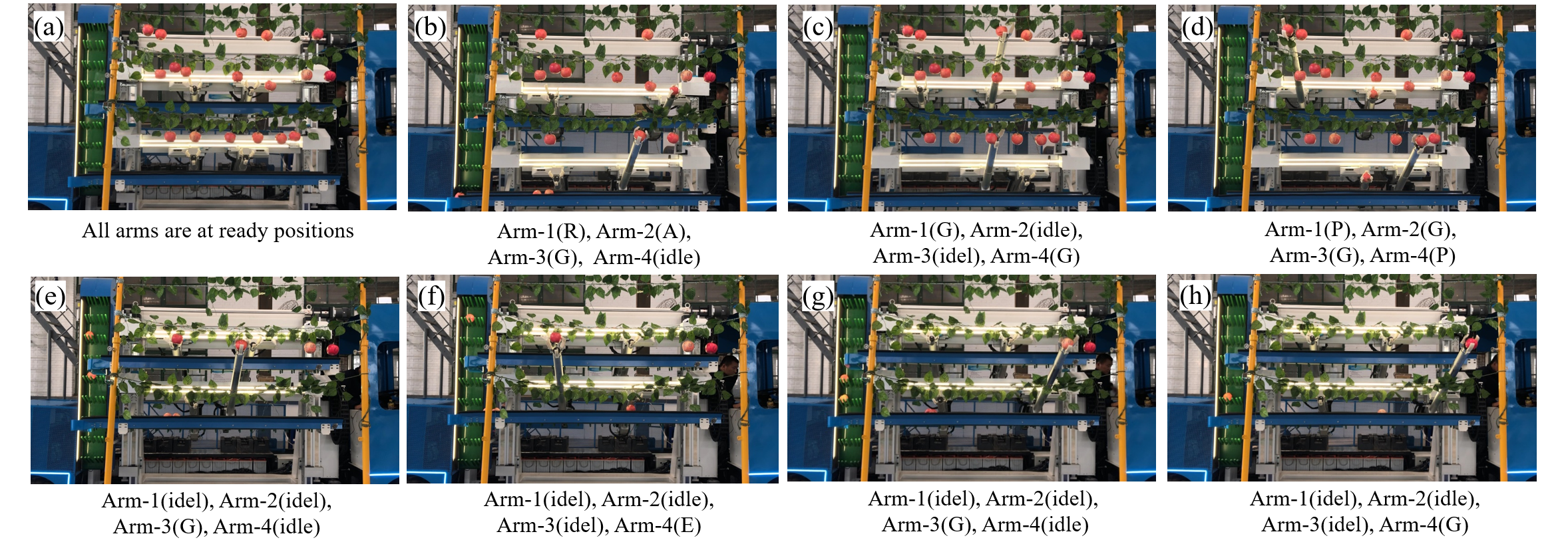}
	\caption{An example of series of state transitions of the multi-arm harvesting robot.}
	\label{fig:multiple-actions}
\end{figure*}
\subsection{Action Definition}
At time $k$, the action of agent-$i$ can be represented by the vector $\mathbf{a}^i_k = {\Psi_k^i, B_{k,1}^i,B_{k,2}^i}$. Here, $\mathbf{a}^i_k \in \mathbf{A}^i$, $\Psi_k^i\in{0,1,\ldots,N}$ represents the number of fruits agent-$i$ will pick, where 0 means no new target and 1$\sim$N means the number of fruits to be picked consistent with $\mathbf{D}_k$. $B_{k,1}^i,B_{k,2}^i\in {0,1}$ represent the action of the left and right arms in the group, respectively. 0 and 1 correspond to the actions of AEG and RP, respectively, and 1 also represents stopping at the placing state.

As mentioned in Section \ref{sec:2}, joint-1 is a coupling joint, and separating the actions of two arms in a group into AEG and RP is necessary to avoid idle time. To better understand the state transitions with different actions, we list 9 state transitions with three action combinations: (0, 1), (1, 0), and (1,1) in Table \ref{tab:transition}. For ease of analysis, we omit $\Psi_k^i$.

For example, if the action of group-U is ${10,0,1}$, it means that group-U will use the left arm to pick No.10 fruit using the A-E-G actions, while the right arm will use the R-P actions to place a fruit at the conveyor or remain in the state of placement. If the action of group-U is ${0,1,1}$, it means that group-U will not pick any fruit, and both the left and right arms will remain in the placement or perform the R-P actions.

\begin{table}[h]
	\centering
	\caption{State transitions of two arms in a group.}
	\label{tab:transition}
	\begin{tabular}{ccccc}
		\toprule[1.5pt]
		Transitions& $s_{k}$     & $s_{k+1}$   & Actions& Description \\
		\midrule[0.8pt]
		$\mathcal{T}_1$ & AEG, RP & RP, RP  & $1,1$& Pause					\\
		$\mathcal{T}_2$ & AEG, RP & RP, AEG & $1,0$& Alternation			\\
		$\mathcal{T}_3$ & AEG, RP & AEG, RP & $0,1$& Non-alternation			\\
		$\mathcal{T}_4$ & RP, AEG & AEG, RP & $0,1$& Alternation				\\
		$\mathcal{T}_5$ & RP, AEG & RP, AEG & $1,0$& Non-alternation		\\
		$\mathcal{T}_6$ & RP, AEG & RP, RP  & $1,1$& Pause					\\
		$\mathcal{T}_7$ & RP, RP  & RP, RP  & $1,1$& Pause					\\
		$\mathcal{T}_8$ & RP, RP  & AEG, RP & $0,1$& Restart					\\
		$\mathcal{T}_9$ & RP, RP  & RP, AEG & $1,0$& Restart					\\
		\bottomrule[1.5pt]
	\end{tabular}
\end{table}

\subsection{Reward Definition}
The rewards in this model consist of three aspects:

(1) The reward of exploration, denoted as $r^i_\text{exp}$. When an agent explores a new target, it receives a positive reward of $+0.05$, which is observed from $\mathbf{D}_t$. If the agent does not explore a new target, then $r^i_\text{exp} = 0$.

(2) The penalty of conflicts, denoted as $r^i_{\text{confl}}$. When two agents collide, they both receive a negative reward of $-0.1$. Otherwise, $r^i_\text{confl} = 0$. Note that we only consider inter-group conflicts, as conflicts within the same group do not affect the time required to reach a target.

(3)Time, which includes the time required for movements and actions. For agent $i$, the time cost from state $s_k$ to state $s_{k+1}$ is determined by:

\begin{equation}\label{eq:timedef}
{^i}t^{k+1}_k  = 
\begin{cases}
{^i}t_\text{AEG}, &\text{if } \mathcal{T}^{k+1}_{k}= \mathcal{T}_2,\mathcal{T}_4,\mathcal{T}_8,\mathcal{T}_9 \\
{^i}t_\text{all}, &\text{if } \mathcal{T}^{k+1}_{k}= \mathcal{T}_3,\mathcal{T}_5\\
{^i}t_\text{idle}, &\text{if } \mathcal{T}^{k+1}_{k}= \mathcal{T}_1,\mathcal{T}_6,\mathcal{T}_7\\
\end{cases}
\end{equation}
where ${^i}t_\text{AEG}$ and ${^i}t_\text{all}$ denote the duration of agent $i$ performing the actions of A-E-G and the complete actions, respectively. ${^i}t_\text{idle}$ denotes the idle time that an agent waits for the other arms and depends on the duration of the other agent's actions when there is inter-agent interference. 
To provide further clarification on the state and action transitions, we present an example in Figure \ref{fig:timesequence}. Specifically, we illustrate $\mathcal{T}_2$ in Figure \ref{fig:timesequence} (a), where the transition from state $k$ to state $k+1$ indicates that ARM-1 has completed its operation and switched to ARM-2, and the time period is denoted by ${^1}t^{k+1}_k = {^1}t_\text{AEG}$. Similarly, $\mathcal{T}_4$ is shown by the transition from state $k+1$ to state $k+2$ in Figure \ref{fig:timesequence} (a), while $\mathcal{T}_8$ is shown by the transition from state $j$ to state $j+1$ in Figure \ref{fig:timesequence} (b).

Next, we illustrate $\mathcal{T}_3$ in Figure \ref{fig:timesequence} (a), where the transition from state $k+2$ to state $k+3$ denotes that ARM-1 has completed its last operation and continued with the next AEG action, calculated by ${^1}t_\text{all} = {^1}t_\text{AEG}+{^1}t_\text{RP}$. The same holds for ARM-2. Meanwhile, $\mathcal{T}_6$ is illustrated by the transition from state $k+2$ to state $k+3$ in Figure \ref{fig:timesequence} (b), which means that ARM-2 has completed its last actions and was in an idle state, with the time cost calculated as the operation time of the other agent, i.e., ${^1}t_\text{idle} ={^2}t^{k+1}_k$. Similarly, $\mathcal{T}_1$ denotes that ARM-1 was in an idle state, having completed its last actions. Lastly, $\mathcal{T}_7$ is illustrated by the transition from state $j-1$ to state $j+1$ in Figure \ref{fig:timesequence} (b), which indicates that both arms were in idle and the duration depended on the operation time of the other agent.
If one agent has no target to visit but the other one has, the agent will maintain the state transitions $\mathcal{T}_7$ to wait for the other agent finishing its tasks. In this way, the two agents can eventually get a same accumulative total reward. Figure \ref{fig:multiple-actions} demonstrates an example of state transitions in a harvesting operation. 

According to the cost of time defined in Eq.\eqref{eq:timedef}, the reward of action $r^i_{\text{time}}$ follows
\begin{equation}\label{eq:rwdtime}
r^i_{\text{time}} = {\alpha}\left(e^{-{^i}t^{k+1}_k}-1\right)
\end{equation}
where $r^i_{\text{time}}$ denotes the reward; $\alpha$ is a scale factor. Eq.\eqref{eq:rwdtime} means that the reward is negative correlation with ${^i}t^{k+1}_k$.

Basing on the above definitions, the reward function is defined by
\begin{equation}
r^i(s_k,a^i_k) = 
\begin{cases}
-50, &\text{if } k=k_\text{max},\\
100, &\text{if all targets picked},\\
r^i_{\text{exp}}+ r^i_{\text{confl}} + r^i_{\text{time}}, &\text{otherwise.}
\end{cases}
\end{equation}

\subsection{Multi-agent Reinforcement Learning}
Collaborative multi-agent reinforcement learning problems require a well-designed collaboration strategy among agents. A direct way is to have a central controller that receives rewards of all agents and determines actions for each agent, called fully centralized. With all agents' information available to the controller, the problem in this work can be modeled as an MDP and solved by single-agent reinforcement learning algorithms. 
In this paper, the two groups of arms are controlled by a central controller, and all information is available. Therefore, the fully centralized strategy is employed in this work, enabling a single-agent RL algorithm to solve the problem with ease.
The multi-arm harvesting Markov game environment involves inter-arm coupling, alternations of inter-group arms, and mechanical limitations, which make the transition dynamics complex. 
Considering the high dimensionality of the state space in this study and the presence of uncertainty in the environment, we therefore use a policy-based learning algorithm, PPO \cite{Schulman2017ProximalPO}. 

PPO is a model-free, online, on-policy, and policy gradient reinforcement learning algorithm, that uses small batches of multiple steps of stochastic gradient descent to optimize the objective function.
In this work, PPO is implemented using stable-baseline 3\cite{raffin2021stable}. The implementation of PPO in this work benefits from a well-designed and efficient framework, which enables the training of complex deep reinforcement learning models with ease.
To train PPO network, the decay factor is set to $\gamma = 0.95$ and the clipping hyperparameter is $\epsilon =0.20$. We use the optimizer Adam\cite{kingma2014adam} with a learning rate of 5e-4 and GAE to estimate advantages, with $\lambda = 0.88$. Training is batch-wise as required by the algorithm, so we use OnPolicyBatchReplay memory. The number of epochs is 8 and the minibatch size is 512.  
To improve the model's generalization ability, we followed a step-by-step training approach. Initially, we trained a simple pre-trained model using an environment with 60 targets, which required 1 million training steps. Next, we further trained the model by exposing it to 10 different environments with varying layouts and target numbers, requiring an additional 1 million training steps. Finally, we further increased the model's adaptability by randomly changing the layout and target numbers at each reset during training, and this resulted in the final policy network model after 100 million training steps.

\section{Simulation and experiments}
\begin{figure}
	\centering
	\includegraphics[width=1\linewidth]{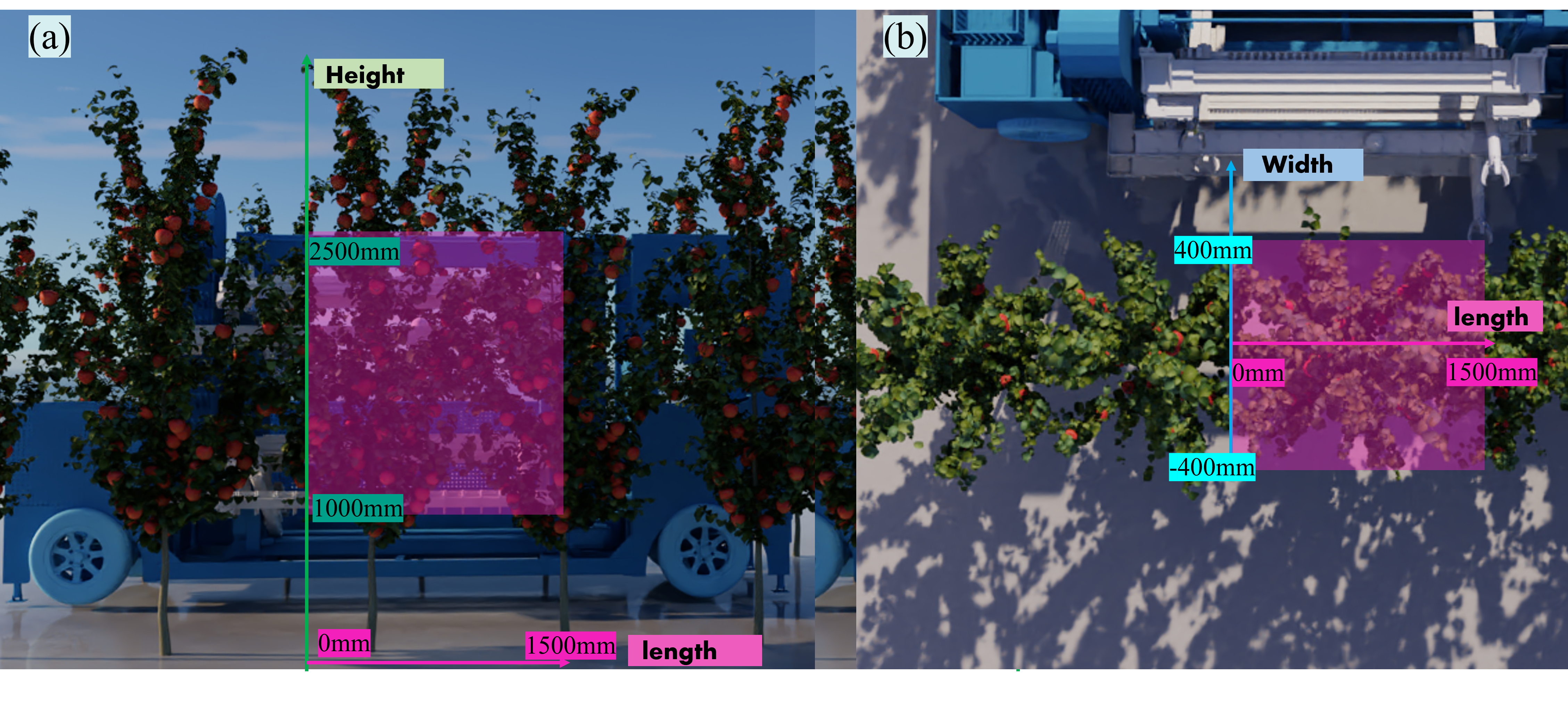}
	\caption{Simulation environment for orchard harvesting powered by NVIDIA Omniverse and schematic diagram of the multi-arm harvester workspace.}
	\label{fig:experiment}
\end{figure}

To demonstrate the effectiveness of the presented method, we conducted verifications by combining simulations and experiments. The simulations were based on a virtual environment powered by NVIDIA Omniverse, while the experiments were performed in an indoor laboratory and an apple orchard, as shown in Fig.\ref{fig:experiment}. 

During the operation, two scenarios were considered in the simulation: one where task execution is regarded as certain, meaning that the fruits are picked sequentially by a robot according to the planning, and another case where the execution is uncertain due to the failures of picking, resulting in some targets undergoing several picking attempts. These two cases have very different effects on the results of task completion.
 
\begin{figure}[h!]
	\centering
	\includegraphics[width=1\linewidth]{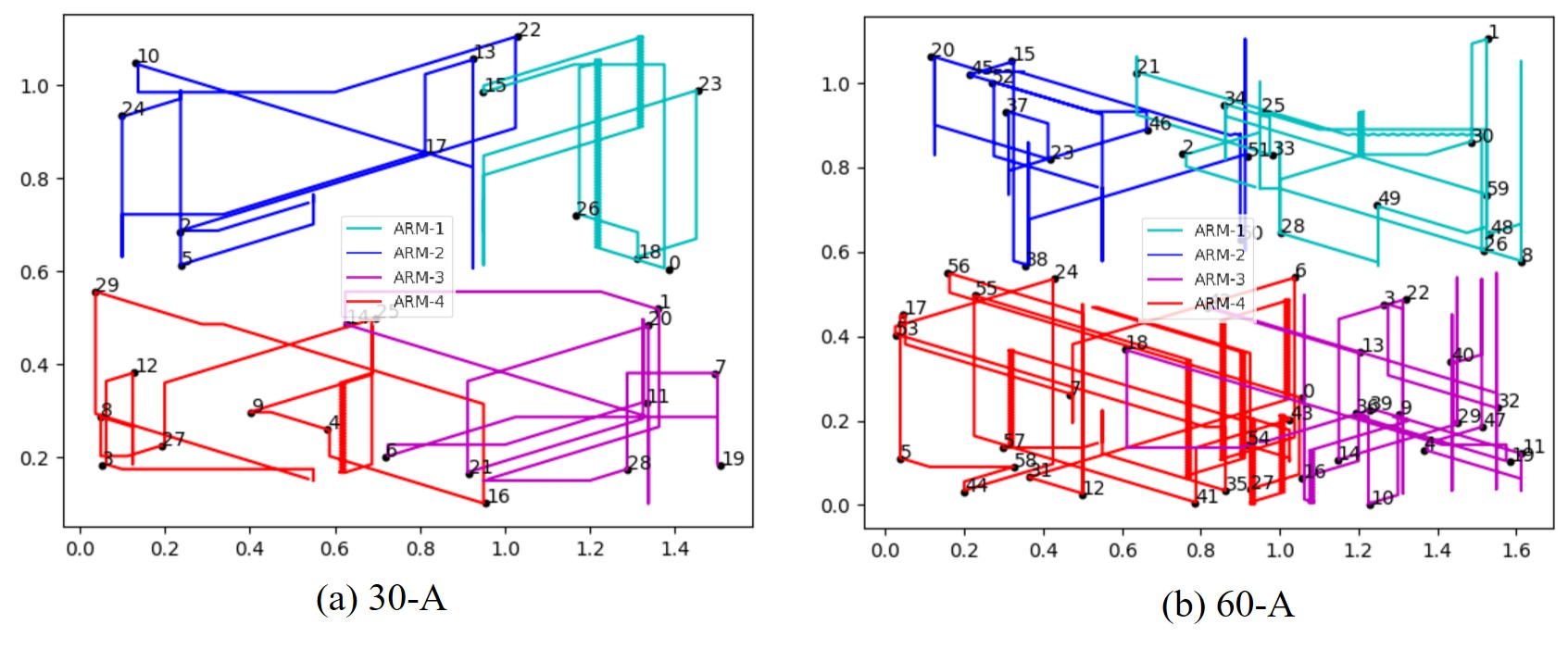}
	\caption{The traversal paths of four arms with the different fruit distributions.}
	\label{fig:targets}
\end{figure}

Considering the varying numbers and distributions of fruits at working sites, we evaluated the performance with 2 different layouts with 30 and 60 fruits. We conducted four groups of tests: 30 fruits and 60 fruits with different distributions A and B by the proposed method and a heuristic-based task planning algorithm proposed in \cite{li2021task}. In each group of tests, we conducted five repetitive tests with the fruit distribution remaining constant in each trial.
In the first case without failed grasp, Table \ref{tab:1} shows that the comparisons between the proposed method and the reference method \cite{li2021task} in five repeated experiments. In terms of average total harvesting time, the reductions were 3.68\% (30-A), 17.56\% (30-B), 5.35\% (60-A), and -1.59\% (60-B), respectively. Although this method and the reference method only showed insignificant advantages in terms of average harvest duration, the presented method significantly reduced the calculation time and improved real-time performance in terms of planning time.
Figure \ref{fig:targets} shows the traversal paths of four arms in the testing groups of 30-A and 60-A.

In the second case with failed grasp, we considered that some fruits could not be successfully grasped in a single attempt and required two or three attempts. When the simulation was conducted using the reference method \cite{li2021task}, it planned only once under the initial conditions, and all the targets traversed were considered as grasped successfully. During the execution, the task was not replanned due to grasping failure. However, when using the proposed method, the task planning was based on the environment and the current state of the robot, and the failed fruits were still attempted. 
During five repeated tests, 7 fruits in 30-A group needed to be picked up twice and 2 fruits were picked up three times. In 30-B group, 8 fruits required 2 attempts and 4 fruits required 3 attempts. In the 60-A group, 10 fruits required 2 attempts and 5 needed 3 attempts; in the 60-B group, 12 fruits needed 2 attempts and 3 required 3 attempts.
Table \ref{tab:2} shows the harvesting experiment results with grasping failure, where the proposed method in this paper increased the total operation time but significantly reduced the remaining number of fruits, improving the quality of the operation.

\begin{table}[h]
	\centering
	\caption{Results of multi-arm harvesting task planning simulation without grasping failure.}\label{tab:1}
	\begin{tabularx}{\linewidth}{XXXXXX}
		\toprule
		Test groups          & Methods & Max.(s) & Min.(s) & Average(s) &Planning Time (s)\\
		\midrule
		\multirow{2}{*}{30-A} & \cite{li2021task}    & 217.60 & 190.00 & 200.78 &  8.25 \\
		& ours     &\textbf{201.40}	&\textbf{177.35}	&\textbf{193.40 }      &\textbf{1.21}\\
		\multirow{2}{*}{30-B} & \cite{li2021task}    & 321.25 & 275.00	&287.43      &8.45 \\
		& ours     & \textbf{241.05} & \textbf{231.30} & \textbf{236.93 }      & \textbf{1.18} \\
		\midrule                    
		\multirow{2}{*}{60-A} & \cite{li2021task}    & 461.35&394.50&435.96    & 15.87\\
		& ours     & \textbf{434.45}&\textbf{387.78}&\textbf{412.64}  &   \textbf{1.24}   \\
		\multirow{2}{*}{60-B} & \cite{li2021task}    & 549.00&\textbf{436.65}&\textbf{490.54 }     &16.42\\
		& ours     & \textbf{536.20}&443.70&498.36     &\textbf{1.14} \\
		\bottomrule                      
	\end{tabularx}
	\begin{tablenotes}
		\footnotesize	\item  Note: In the table, max., min. and average denotes the maximum, the minimum and the average of harvesting operation duration among five repetitive trials. Planning time means the time cost to calculate the planning results.
	\end{tablenotes}
\end{table}

\begin{figure*}
	\centering
	\includegraphics[width=\linewidth]{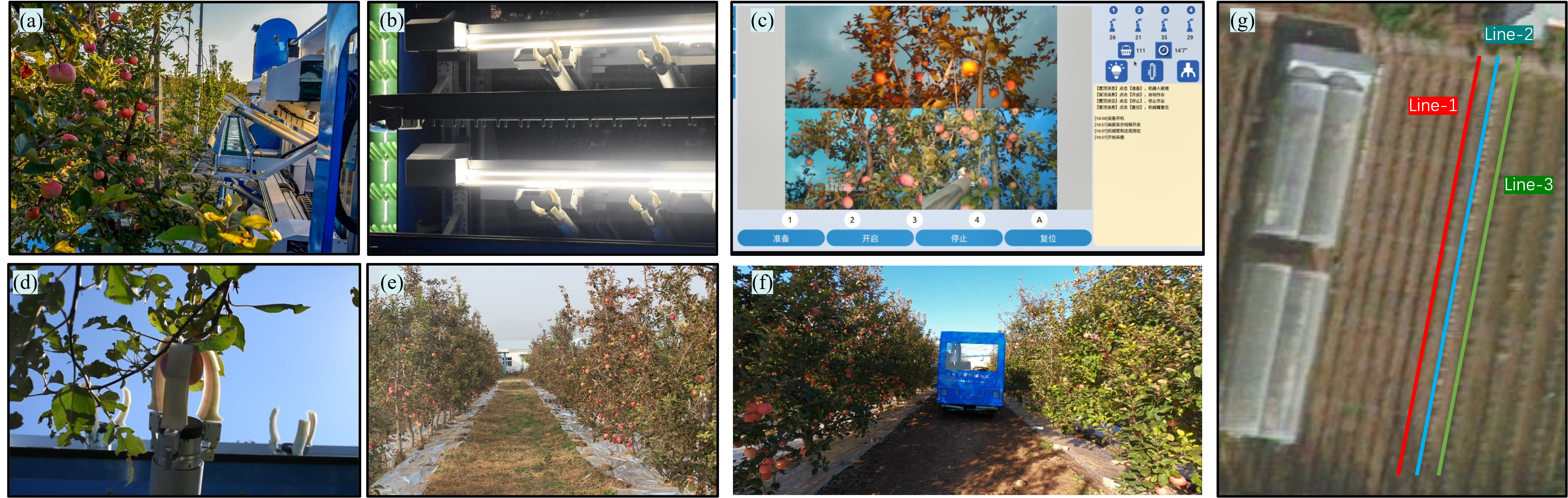}
	\caption{The four-arm harvesting robot and the apple orchard in the experiments. (a) The multi-arm harvesting robot at a working site; (b) The illuminating system is working; (c) The software platform of the robot; (d) A gripper is grasping an apple target; (e) A pathway between rows of fruit trees; (f) The robot is moving along a pathway; (g) The tree lines used in the orchard experiments.}
	\label{fig:orchardexperiment}
\end{figure*}

\begin{table}
	\centering
	\caption{Results of the multi-arm harvesting task planning simulation with grasping failures.}\label{tab:2}
	\begin{tabular}{cccccccc}
		\toprule
		\multirow{2}{*}{Tests} & \multirow{2}{*}{Methods} & \multirow{2}{*}{Average Time (s)} & \multicolumn{5}{c}{Remaining fruits}  \\
		&                           &                         & \textrm{I}  & \textrm{II}  & \textrm{III}  & \textrm{IV}  & \textrm{V}                    \\
		\midrule                        
		\multirow{2}{*}{30-A}   & \cite{li2021task}                     & 159.27                  & 9  & 9  & 9  & 9  & 9                    \\
		& ours                      & 237.70                  & 1  & 0  & 1  & 0  & 0                    \\
		\multirow{2}{*}{30-B}   & \cite{li2021task}                     & 164.76                  & 12  & 12  & 12  & 12  & 12                    \\
		& ours                      & 259.54                  & 1  & 3  & 2  & 1  & 1                    \\
		\midrule
		\multirow{2}{*}{60-A}   & \cite{li2021task}                     & 339.03                  & 15 & 15 & 15 & 15 & 15                   \\
		& ours                      & 434.86                  & 4  & 3  & 1  & 1  & 4                    \\
		\multirow{2}{*}{60-B}   & \cite{li2021task}                     & 329.32                  & 15 & 15  & 15 & 15 & 15                   \\
		& ours                      & 479.26                  & 5  & 4  & 4  & 3  & 5                    \\
		\bottomrule 
	\end{tabular}
\end{table}
In practical orchard experiments, we selected three rows of apple trees with similar conditions of canopy and quantities of fruits in a standard apple orchard, as shown in Figure \ref{fig:orchardexperiment}(b) and (c), and conducted orchard experiments. In each row of trees, about 30 working sites were selected to conduct three testing groups: the groups of line-1, line-2 and line-3 employing random traversal method, the reference method \cite{li2021task}, and the proposed method, respectively. The random planning method refers to not using any task planning method, but randomly assigning the targets for each arm.
\begin{table*}[t]
	\centering
	\caption{Harvesting performance in full field trails in an apple orchard}\label{tab:3}
	\begin{tabularx}{\textwidth}{XXXXXXXXX}
		\toprule[1.5pt]
		Groups&  Working sites &  Apple & Fruits Reachable(\%) &  Picked &  Picked w.r.t. reachable &  Lost &  Average Cycle-time (s)&Methods\\
		\midrule
		Line-1 & 30 & 453 & 186(41.06\%) & 149(32.89\%) & 80.11\% & 37 & 7&Random \\
		Line-2 & 35 & 420 & 176(41.90\%) & 120(28.57\%) & 68.18\% & 56 & 6.1&\cite{li2021task} \\
		Line-3 & 28 & 382 & 145(37.96\%) & 115(30.10\%) & 79.31\% & 30 & 5.8 &Ours \\
		\bottomrule
	\end{tabularx}
\end{table*}

As shown in Table \ref{tab:3}, the proportion of reachable fruits is not high w.r.t. the total, about 40\%, which is because the lowest fruit that this robot can pick is 1.5m above the ground, and a considerable number of fruits are below 1.5m in the experiments; in addition, some fruits are blocked by branches and difficult to approach. The experimental results in Table \ref{tab:3} show that: 1) the experimental groups using task planning algorithms, i.e., the reference method and the proposed method, have an average operating time at each site significantly lower than the random traversal method, indicating that task planning can effectively reduce operation time. 2) The index of picked w.r.t. reachable of the comparative method is lower, only 68.18\% , which is significantly lower than the other two. It indicates that the offline planning method lacks dynamic updating of tasks and makes it difficult to handle the situation of failed fruit picking. 3) The average cycle-time of the proposed method is significantly lower than the random traversal method, and slightly lower than the comparative method. In this experiment, limited by the reachable range, the actual number of fruit targets per working site is about 15 on average, and when the number of fruits is small, the superiority of this method in terms of operation time cannot be fully demonstrated; however, this method demonstrates its superiority in terms of harvest rate, with significantly fewer lost fruits than the comparative method. Overall, the orchard experiment shows that the four-arm picking robot using the method proposed in this paper has a successful picking rate of about 79.31\% and an average picking time of 5.8s per fruit.

\section{Conclusions and future work}\label{sec con}
This study presents a novel task planning strategy for a four-arm harvesting robot in apple orchards to enhance operational efficiency. To achieve effective collaboration among the arms, various constraints of the robot, including inner-arm conflicts, joint coupling, mechanical limitations, and grasping failures, have been considered. A Markov game framework has been employed to formulate the cooperative harvesting problem for the four arms. To facilitate real-time decision-making for the multiple arms while minimizing computational complexity, a MARL-based task planning method has been proposed.
The efficacy of the proposed method has been verified through simulations employing four distinct fruit layouts. The results show that the proposed method is comparable to heuristic algorithms in terms of planning effectiveness, while demonstrating significantly improved planning efficiency, particularly with a large number of fruits. In contrast to heuristic methods, which required approximately 8s (30 fruits) or 16s (60 fruits), the task planning solution could be obtained in about 1s. Additionally, the proposed method exhibited satisfactory harvesting completion rates and real-time planning performance, accounting for the possibility of failed grasp.
Experimental results from the orchard study demonstrate the practical applicability of the proposed method, with fewer lost fruits and shorter average harvesting cycles being observed.

In the experimental evaluation, it was observed that the robot's reachable working range was limited, resulting in a fruit accessibility rate of approximately 40\%. To improve the efficiency of the robotic harvesting process, future efforts are needed to optimize the robot's structural design to increase accessibility to fruits located in the lower part of the canopy. This limitation will be the focus of our future research endeavors.
\addtolength{\textheight}{-6cm}

\end{document}